\documentclass[10pt,twocolumn,letterpaper]{article}

\usepackage{cvpr}              

\usepackage{graphicx}
\usepackage{amsmath}
\usepackage{amssymb}
\usepackage{pifont}
\usepackage{booktabs}
\usepackage{multirow} 
\usepackage{tabularx} 
\usepackage[accsupp]{axessibility}  

\usepackage[pagebackref,breaklinks,colorlinks]{hyperref}

\usepackage[capitalize]{cleveref}
\crefname{section}{Sec.}{Secs.}
\Crefname{section}{Section}{Sections}
\Crefname{table}{Table}{Tables}
\crefname{table}{Tab.}{Tabs.}

\def\cvprPaperID{4940} 
\def\confName{CVPR}
\def\confYear{2023}

\begin{document}

\title{FocalDreamer: Text-driven 3D Editing via Focal-fusion Assembly}
\author{Yuhan Li\textsuperscript{1}\hspace{2mm}
Yishun Dou\textsuperscript{2}\hspace{2mm}
Yue Shi\textsuperscript{1}\hspace{2mm}
Yu Lei\textsuperscript{1}\hspace{2mm}
Xuanhong Chen\textsuperscript{1}\hspace{2mm}
Yi Zhang\textsuperscript{1}\hspace{2mm}
Peng Zhou\textsuperscript{1}\hspace{2mm}
Bingbing Ni\textsuperscript{1$\dagger$}\\
\textsuperscript{1}Shanghai Jiao Tong University, Shanghai 200240, China \qquad \textsuperscript{2}Huawei \\
{\tt\small \{melodious, nibingbing\}@sjtu.edu.cn}\\
{\small \url{https://focaldreamer.github.io}}
}

\maketitle

\begin{abstract}
While text-3D editing has made significant strides in leveraging score distillation sampling, emerging approaches still fall short in delivering separable, precise and consistent outcomes that are vital to content creation. In response, we introduce FocalDreamer, a framework that merges base shape with editable parts according to text prompts for fine-grained editing within desired regions. Specifically, equipped with geometry union and dual-path rendering, FocalDreamer assembles independent 3D parts into a complete object, tailored for convenient instance reuse and part-wise control. We propose geometric focal loss and style consistency regularization, which encourage focal fusion and congruent overall appearance. Furthermore, FocalDreamer generates high-fidelity geometry and PBR textures which are compatible with widely-used graphics engines. Extensive experiments have highlighted the superior editing capabilities of FocalDreamer in both quantitative and qualitative evaluations.
\end{abstract}

\newcommand{\customfootnotetext}[2]{{
  \renewcommand{\thefootnote}{#1}
  \footnotetext[0]{#2}}}
\customfootnotetext{${\dagger}$}{Corresponding author: Bingbing Ni.}

\section{Introduction}

Art reflects the figments of human imagination and creativity. Recently, the rapid development of neural generative model~\cite{ho2020ddpm, dhariwal2021adm} has significantly lowered the barriers for humans to engage in artistic creation with just a few words. However, these black-box models also deprive humans of a significant portion of control, which means the generated results aren't often aligned with expectations. In this work, we take a step towards precise editing for 3D creation, enabling neural networks to naturally expand user's intentions, rather than controlling the entire generative process.

In the realms of animation, gaming, and the recent advance of virtual augmented reality, 3D models and scenes are commonly constructed as an assembly of semantically distinct base parts, which support the practice of rendering multiple copies of the same part across scenes with different transform matrices, called \emph{geometry instancing} or \emph{instance reuse} (Fig.~\ref{fig: motivation}). We believe that an ideal 3D editing workflow should possess the following good properties:

\begin{figure}[t]
\centering
\includegraphics[width=0.95\columnwidth]{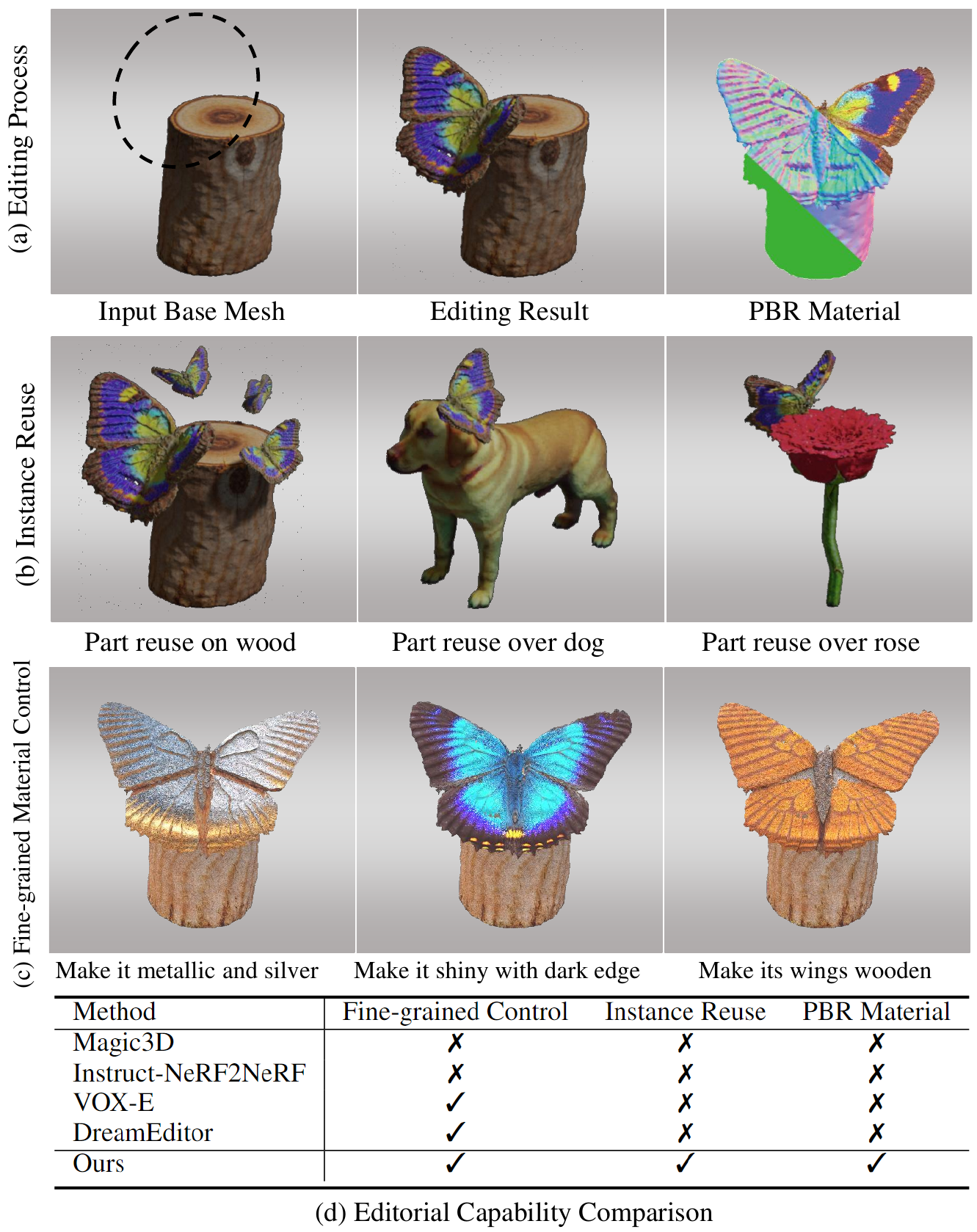}
\caption{Given the prompt ``a butterfly over a tree stump", our method delivers high-fidelity geometry and photorealistic appearance using PBR materials. Lines (b-c) showcase instance reuse and part-wise material control, underscoring FocalDreamer's capability for separable and precise edits.}
\label{fig: motivation}
\end{figure}

\begin{figure*}[t]
\centering
\includegraphics[width=2.1\columnwidth]{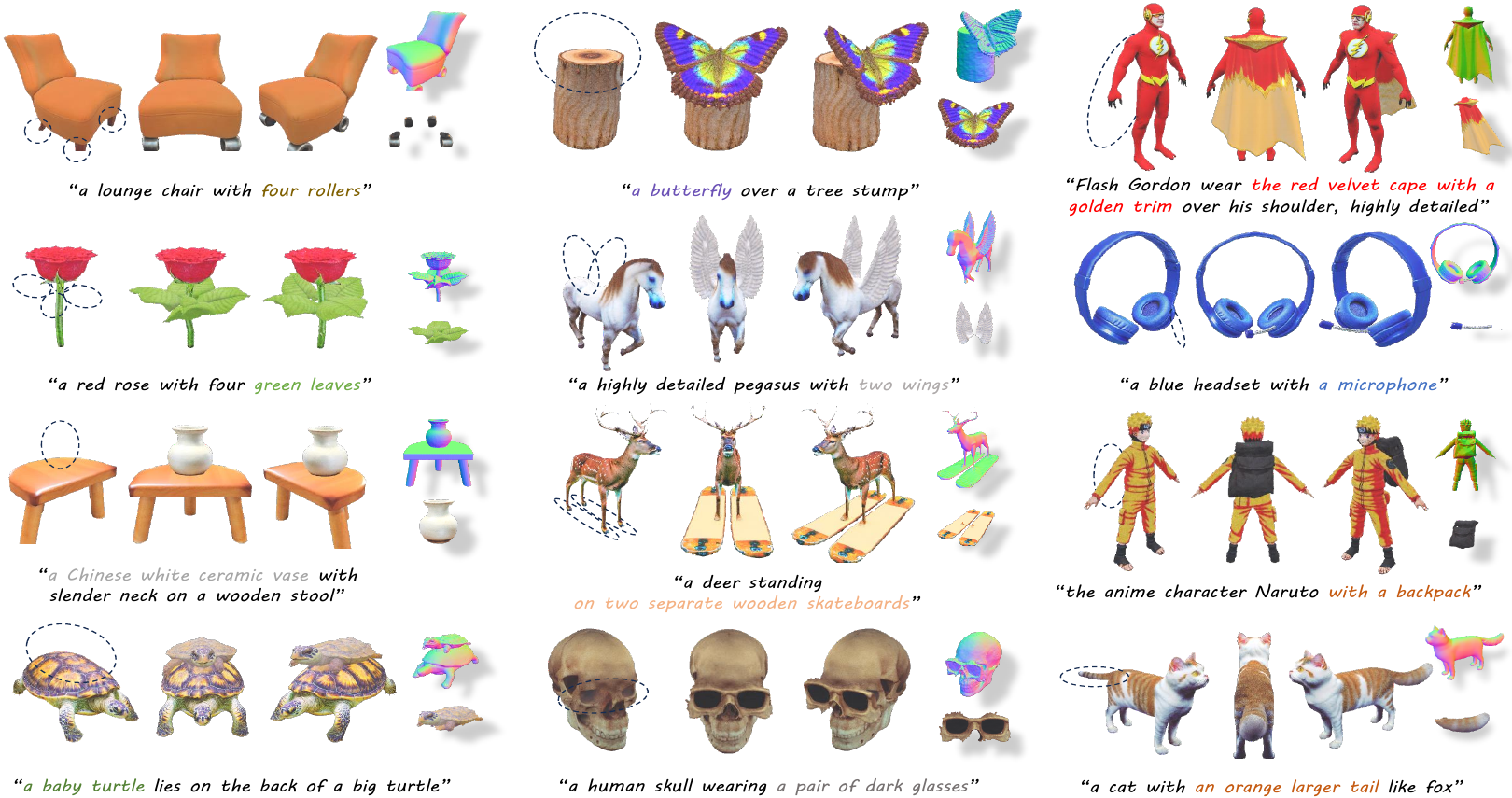} 
\caption{FocalDreamer can generate meticulously detailed and photo-realistic 3D editing. The left column displays base meshes with focal regions. The three right columns showcase edited overall appearance, assembled geometry, and editable part.}
\label{fig: main}
\end{figure*}

\begin{itemize}
\item \textbf{Separable.} Given a base shape, it should produce structurally \textit{separate parts}~\cite{li2020assembly} facilitating for instance reuse and part-wise post-processing, grounded in widespread understanding. 
\item \textbf{Precise.} It should provide \textit{fine-grained} and \textit{local} editing, enabling precise control in the desired area~\cite{zhuang2023dreameditor}, while maintaining other regions untouched.
\item \textbf{Consistent.} After the editing process, the resultant shape should respect the characteristics of the source shape in \textit{harmonious appearance}~\cite{xie2023smartbrush}, while visually adhering to the text specifications.
\end{itemize}

Emerging approaches in text-3D editing have achieved noteworthy development, yet they often fall short in delivering separable, precise, and consistent outcomes that are vital to content creation. Some approaches~\cite{lin2023magic3d,haque2023instruct} struggle to pinpoint the focused local regions, leading to undesired alterations to the base shape. Others~\cite{sella2023vox,zhuang2023dreameditor} overlook the stylistic consistency of the 3D edited portions. Furthermore, nearly all past methods directly modify the base shape, neglecting the need for \emph{instance reuse} and \emph{part-wise control} (\textit{i.e.}, enabling fine-grained edits to individual parts of a complete object). Moreover, their coupling of geometry and textures has compromised the quality of the edits.

\begin{figure*}[t]
\centering
\includegraphics[width=2.1\columnwidth]{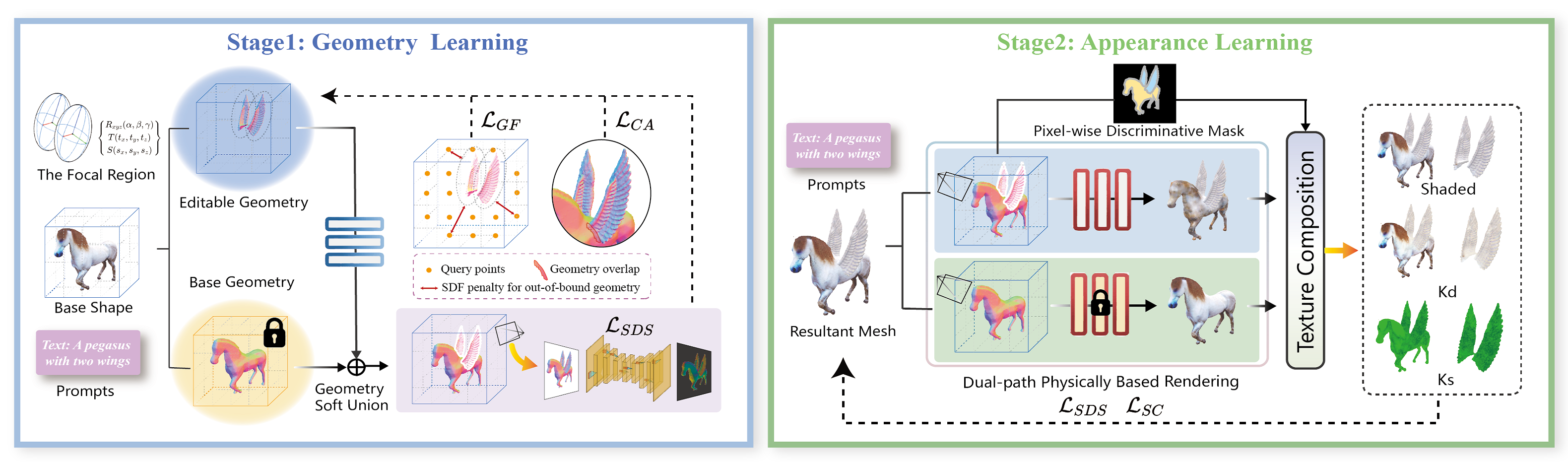} %
\caption{An overview of FocalDreamer. (a) During geometry learning, given a base shape, we first initialize an ellipsoid as editable geometry within each focal region. Then we render the normal map of merged shape as shape encoding of pre-trained T2I models, to optimize the editable geometry according to prompts. (b) During appearance learning, resultant shape is rendered in a dual-path manner with base and editable textures. The outcomes are then blended by Pixel-wise Discriminative Mask for a unified appearance. (c) Several regularizations are introduced to improve the editing quality, including $\mathcal{L}_{GF}$, $\mathcal{L}_{CA}$, and $\mathcal{L}_{SC}$.}
\label{fig: framework}
\end{figure*}

We introduce the following key contributions to meet our outlined criteria:
(1) \textbf{Separable}: we propose FocalDreamer, a user-friendly framework that permits intuitive object modifications using text prompts and a rough focal region for the intended edits. Instead of direct modifications to the \textbf{base shape} (\textit{e.g.}, the \textit{horse} in Fig.~\ref{fig: framework}), a novel \textbf{editable part} (\textit{wings} in Fig.~\ref{fig: framework}) is generated in the focal region, facilitating instance reuse and precise control. Equipped with geometry union and dual-path rendering, this part is merged with base mesh into a semantically unified shape in a lossless and differentiable manner, then optimized using a powerful text-to-image model to align the prompts and shapes. Furthermore, our decoupled learning of geometry and appearance yields detailed geometry and PBR textures, ensuring compatibility with prominent graphics engines. 
(2) \textbf{Precise}: Users delineate one or several ellipsoid focal regions, in which a spherical editable part initializes, acting as a smooth prior for the geometry network. The geometric focal loss is also introduced, discouraging edits beyond specified regions. 
(3) \textbf{Consistent}: a smooth, coherent surface is essential in certain scenes. Hence, a soft geometry union operator and a style consistency regularization are proposed to ensure a seamless geometric transition and stylistically consistent texture between the learnable part and base shape.

To our knowledge, this is the first component-based editing method with separate learnable parts. Rich experiments and detailed ablation studies highlight the superior editing capabilities of our approach, as shown in Fig.~\ref{fig: main}.

\section{Related work}

\noindent \textbf{Text-guided Image Generation and Editing.}
Significant progress in Text-to-Image (T2I) generation with diffusion models~\cite{ho2020ddpm} is witnessed in recent years. More recently, with the availability of scalable generator architectures and extremely large-scale image-text paired datasets, they've demonstrated impressive performance in high-fidelity and flexible image synthesis~\cite{rombach2022ldm}. Due to their comprehension of complex concepts, diffusion models are also amicable for various editing tasks, such as image inpainting~\cite{lugmayr2022repaint}, image blending~\cite{avrahami2022blended}, image stylization~\cite{zhang2023inversion}. The most relevant field to us among those is inpainting, which provides flexible control of the inpainted content, and a mask to constrain the shape of the inpainted object. SmartBrush~\cite{xie2023smartbrush} introduces a precision factor into the masks for multiple-grained controls on inpainting regions. 

\noindent \textbf{Text-to-3D Content Generation.}Driven by the aspiration to produce high-fidelity 3D content using semantic inputs like text prompts, the field of text-to-3D has garnered a significant boost in recent years~\cite{poole2022dreamfusion}. Earlier approaches either align shapes and images in the latent space by CLIP supervision~\cite{radford2021clip} to generate 3D geometries~\cite{mohammad2022clipmesh} or synthesize new perspectives~\cite{jain2022zero}, or they train text-conditioned 3D generative models from the ground up~\cite{li2023generalized}. DreamFusion~\cite{poole2022dreamfusion} first employs large-scale T2I models with a combination of score distillation sampling to distill the prior, and achieves impressive results. Magic3D~\cite{lin2023magic3d} further improved the quality and performance of generated 3D shapes with a 2-step pipeline. Fantasia3D~\cite{chen2023fantasia3d} and TextMesh~\cite{tsalicoglou2023textmesh} modify the representation of geometry to extract detailed mesh and photorealistic rendering.
However, all these methods present semantic misalignment between the local content and global text description when editing, leaning towards distorted background and inconsistent results. Therefore, we proposed a dual-branch framework with fine-grained editing on ideal regions. 

\noindent \textbf{3D Content Editing.}Semantic-driven 3D scene editing is a mulch harder task compared with 2D photo editing because of the high demand for multi-view consistency, the scarcity of paired 3D data and its entangled geometry and appearance. Previous approaches either rely on laborious annotation~\cite{kania2022conerf, yang2022neumesh, yuan2022nerfedit, yang2021learning}, only support object deformation or translation~\cite{tschernezki2022neural, kobayashi2022decomposing, gao2022nerf}, or only perform global style transfer~\cite{chen2022upst, chiang2022stylizing, fan2022unified, huang2022stylizednerf, zhang2022arf} without strong semantic meaning. Recently, thanks to the development of score distillation sampling technique, text-guided editing has emerged as a promising direction with great potential. SKED~\cite{mikaeili2023sked} possesses the capability to edit 3D scenes by utilizing precise multi-view sketches. Latent-NeRF~\cite{metzer2023latentnerf} and Fantasia3D~\cite{chen2023fantasia3d} realize sketch-shape guidance by relaxed geometric constraints on the sketches' surface. Instruct-NeRF2NeRF~\cite{haque2023instruct} can edit an existing NeRF scene by iterative dataset update. However, it manipulates the entire space, and the preservation of undesired regions is absent. Vox-E~\cite{sella2023vox} allows local edits on an existing NeRF, but it suffers from subpar editing quality and noticeable noise as shown in Section~\ref{sec: experiment}, because of coupling geometry and textures. Most related to our work, DreamEditor~\cite{zhuang2023dreameditor} locally edits a mesh-based neural field. However, it doesn’t achieve separable editing which is vital for instance reuse and part-wise control. Moreover, DreamEditor cannot change the number of vertices, supporting only minor shape insertion and replacement of objects of the same type (\textit{e.g.}, replacing a horse with a deer). In contrast, our work not only brings about highly reasonable and noticeable geometric changes but also generates realistic appearances.


\section{Method}

As illustrated in Fig.~\ref{fig: framework}, a complete object is conceptualized as a composition of base shape and learnable parts, wherein both of them possess their own geometry and texture, tailored for convenient instance reuse and part-wise control. Furthermore, a two-stage training strategy is adopted to sequentially learn the geometry and texture of the editable shape, to avoid the potential interference that can occur when geometry and texture learning are intertwined. For instance, in the case of \textit{zebra} modeling, geometric protrusions might be learned instead of the desired black stripes. Such a disentangled learning approach not only stabilizes the training process but also yields high-fidelity geometry and textures, especially when compared to mainstream text-to-3D techniques~\cite{poole2022dreamfusion,sella2023vox}. The details of above two stages are explained in the following sections.

\subsection{Preliminary}
\noindent \textbf{Score Distillation Sampling.}
Score distillation sampling (SDS) is a way to distill the priors hidden in large T2I models for 3D generation proposed by DreamFusion~\cite{poole2022dreamfusion}. DreamFusion represents 3D scenes as a series of learnable parameters $\theta$. Utilizing a differentiable renderer, it converts the 3D scenes into 2D image sets $x$. Subsequently, it employs large-scale models $\phi$ to optimize the parameters of the 3D scenes with a score function. The SDS loss is calculated as the per-pixel gradient as follows:
\begin{equation}
\small \nabla _\theta \mathcal{L} _{SDS}(\phi ,x)= \mathbb{E} _{t,\epsilon }\left [ w(t)(\hat{\epsilon} _{\phi}(z_t;y,t)-\epsilon )\frac{\partial x}{\partial \theta}  \right ] ,
\end{equation}
where $w(t)$ controls the weight of SDS guidance depending on noise level $t$. $\hat{\epsilon} _{\phi}(z_t;y,t)$ and $\epsilon $ are the predicted noise and actual noise, respectively. $y$ is the condition.

\noindent \textbf{DMTet.} DMTet~\cite{munkberg2022dmtet} is a hybrid representation that has two components, \textit{i.e.}, a deformable tetrahedral grid and a differentiable Marching Tetrahedral (MT) layer. The Signed Distance Function (SDF) values and the position offsets of deformable tetrahedral vertices are learnable, followed by the MT layer to extract meshes. It is capable of generating high-resolution 3D shapes due to its high memory efficiency.

\subsection{Geometry Editing}

\noindent \textbf{Focal Region.} The starting point of our algorithm is a base shape ($\Psi _b$ for geometry and $\Gamma _b$ for texture) to be edited, which can be the reconstruction from images, crafted shapes by artists~\cite{munkberg2022dmtet}, and even the novel shapes from the generative method~\cite{chen2023fantasia3d}. Then the base model is modified by compositing with a new learnable part according to prompts. To offer more precise control over the generation process, users are requested to select one or multiple ellipsoid areas (depending on the editing needs) as focal/target regions. Each focal region $\Omega '$ is deformed from a standard sphere $\Omega $ by an affine transformation with $9$ degrees of freedom (DOF), $3$ DOF for stretching, $3$ DOF for rotation, and $3$ DOF for translation along the \{X, Y, Z\}\text{-axis}:
\begin{equation}
\small \Omega '=R_{xyz}(\alpha, \beta, \gamma)\cdot T(t_x, t_y, t_z)\cdot S(s_x, s_y, s_z)\cdot \Omega .
\end{equation}
The selection of the focal region doesn't require exact precision for it merely serves as a rough expression of the regional prior from user intent. Our model will optimally generate geometry driven by the text input. Furthermore, we initialize ellipsoids within specified regions, offering a smooth prior that enhances the stability of the geometric modeling.

\noindent \textbf{Geometry Learning and Fusion.} We adopt DMTet as our 3D scene representation optimized by the prior knowledge distilled from pre-train T2I model. More specifically, keeping the base shape $\Psi _b(v_i)$ frozen, we parameterize the SDF values (inner is positive) of editable parts using MLP $\Psi _e(v_i)$ for each vertex $v_i$ within the tetrahedral grid. Subsequently, a soft geometry union~\cite{union} is performed between $\Psi _b(v_i)$ and $\Psi _e(v_i)$, resulting in $\Psi_u(v_i)$ for a smooth junction:
\begin{align}
\small \Psi _u(v_i) = \max \left \{\Psi _b(v_i), \Psi _e(v_i) \right \}  + \frac{0.1\times h^2}{k}, \\
\small \text{where } h = \max \left \{ (k-|\Psi _b(v_i)-\Psi _e(v_i)|) , 0 \right \}, 
\label{eq: union}
\end{align}
where $k$ determines the extent of the soft merge and is set to $0.15$ by default. After geometry fusion, a differentiable MT layer transforms $\Psi _u(v_i)$ and the vertex offset $\Delta v_i$ into a triangular surface mesh $\mathcal{M}$. Finally, the rendered normal map $n$ and the object mask $o$ extracted from the mesh$\mathcal{M}$ are fed into pre-trained T2I models with SDS loss to update $\Psi _e$:
\begin{equation}
\small \nabla _{\Psi_e} \mathcal{L} _{SDS}(\phi ,\tilde{n} )= \mathbb{E} _{t,\epsilon }\left [ w(t)(\hat{\epsilon} _{\phi}(z_t^{\tilde{n}};y,t)-\epsilon )\frac{\partial \tilde{n}}{\partial \Psi}\frac{\partial z^{\tilde{n}}}{\partial \tilde{n}}  \right ] ,
\end{equation}
where $\phi $ parameterized pre-train T2I model, $ \tilde{n}$ represents the augmentation of $n$ concatenated with $o$, $ z^{\tilde{n}}$ is latent encoding of $ \tilde{n}$. We observed using normal map $n$ promotes the expression of geometric details and training stability~\cite{chen2023fantasia3d}. This improvement from $n$ is partly attributed to disentangling the geometry from the intertwinement of texture, and its sufficient expressiveness to depict complex geometric details.

\noindent \textbf{Geometric Concentration.} One of the main criteria for a proficient 3D editing algorithm is its ability to retain the geometry and color of the base object throughout the editing process. However, the aforementioned pipeline cannot ensure locality in editing. We have observed global changes and a loss of characteristics from the base shape (Fig.~\ref{fig: ablation}). To counteract it, we introduce distance-aware \textbf{geometric focal loss} $\mathcal{L}_{GF}$. During each iteration, a certain number of points $p_i\in \mathbb{R}^3$ are sampled outside the user-specified focal region $\Omega '$, with their SDF values $\Psi _e(p_i)$ and their distances $d_i$ to the focal region $\Omega '$. The objective of $\mathcal{L}_{GF}$ is punishing the editable shape when it produces topological structures ($\Psi_e(p_i)>0 $) outside $\Omega '$. Moreover, the closer $p_i$ is to the target region, the less the penalty, for this distance-aware setting permits geometry to overrun beyond the rough focal region slightly. The \textbf{geometric focal loss} is defined as:

\begin{equation}
\small \mathcal{L}_{GF}=\mathbb{E}_{p_i \notin \Omega '} \left [ (1-e^\frac{-d_i^2}{\sigma_1} )\cdot \tanh (\frac{\max \left \{ \Psi _e(p_i)+\xi , 0 \right \}}{\sigma_2}) \right ] ,
\end{equation}
where $\sigma_1=0.05$ and $\sigma_2=0.01$ control how sensitive the loss is, \textit{i.e.}, lower $\sigma $ values tighten the constraint on the optimization such that only the editable region is modified strictly. The hyperparameter $\xi$ is a small positive threshold to prevent topological structures from minor positive SDF values. For computational efficiency, we sample query points on the tetrahedral vertex $v_i$, and pre-compute their distance $d_i$ to $\Omega '$ before the geometry generation process begins.

\noindent \textbf{Collision Avoidance.} Another essential criterion is to respect the purity of the editing results, \emph{i.e.}, the editable shape should not overlap with the base shape, as they are semantically independent and distinct parts. We enforce it by penalizing the query points $p_i$ that reside both within the learnable shape and the base shape with the \textbf{collision avoidance loss}:

\begin{equation}
\small \mathcal{L}_{CA}=\mathbb{E}_{p_i} \left [ \max \left \{ \Psi _b(p_i), 0 \right \} \cdot \max\left \{ \Psi _e(p_i), 0 \right \}  \right ].
\end{equation}
Intuitively, this reduces the likelihood of overlap between the editable shape and the original mesh, resulting in cleaner editing outcomes. For computational efficiency, we sample query points at $v_i$ as the same as geometric focal Loss.

\subsection{Appearance Editing}

\noindent \textbf{Dual-path Physically Based Rendering.}\label{sec: texture} After the optimization of the geometry network, the resultant mesh $\mathcal{M}$ is obtained from the soft fusion and MT layer. Following Physically Based Rendering (PBR) material model, we use hash-grid-based texture neural fields $\Gamma$ for $\mathcal{M}$ to produce the diffuse term $k_d$, the roughness and metallic term $k_{rm}$, and the normal term $k_n$ as $(k_d, k_{rm},k_n)=\Gamma (p_i)$. In order to retain the appearance of the base shape untouched, a naive and straightforward idea would be to initialize the learnable texture neural fields $\Gamma _e$ with the base texture fields $\Gamma _b$ derived from the base shape reconstruction, then the entire shape's appearance is modeled by $\Gamma _e$ exclusively. However, this simple pipeline has two shortcomings: 1) As the number of iterations increases, it suffers from sub-optimal convergence and loss of the original material (in Fig.~\ref{fig: ablation}). In essence, the texture of the base shape isn't adequately retained due to the overly strong knowledge supervision from T2I models. 2) Although learnable parts have independent semantics, such as ``the wings", their texture cannot be extracted alone. This impediment makes the reuse and driving of materials for these editable parts unfeasible.

 To tackle this issue, we re-design the rendering pipeline in a dual-path manner. Central to this redesign is a Pixel-wise Discriminative Mask (PDM) generated in the rasterization process, which discerns whether each pixel comes from the face of the base mesh or the editable mesh. As depicted in Fig~\ref{fig: framework}, throughout the dual-path rendering process, both parts are rendered based on their own neural texture fields, and the outcomes are then blended by PDM, which is called texture composition, culminating in a unified merged view. Similarly, the merged view is inputted into the T2I model for texture optimization with SDS loss. By truncating the gradient towards $\Gamma _b$, the texture of the base shape is precisely preserved, while the editable shape has its independent trainable texture $\Gamma _e$. As is clear, dual-path rendering balances the preservation of the base shape structure with flexible part-wise control, as well as the seamless integration of the base part and editable part.

\noindent \textbf{Style Consistency.} 
In some instances, local changes are anticipated to be realized seamlessly, as well as in a harmoniously coordinated style, as shown in Fig.~\ref{fig: ablation}. This problem is modeled as follows: let $\mathcal{M}_{e} \in \mathbb{R}^3$ be a closed subspace to represent the editable parts with boundary $\partial \mathcal{M}_{e}$. Let $f^*$ be a known mapping function defined over $\mathbb{R}^3$ minus the interior of $\mathcal{M}_{e}$ to be preserved, and let $f$ be the unknown function defined over the interior of $\mathcal{M}_{e}$. A classical interpolant $f$ is defined as the solution of the minimization problem in image inpainting~\cite{perez2003poisson}:
\begin{equation}
\small \underset{f}{\min} \iiint_{\mathcal{M}_{e}}\left | \nabla f \right | ^2 \text{ with } f|_{\partial \mathcal{M}_{e}}=f^*|_{\partial \mathcal{M}_{e}} .
\end{equation}

We propose two consistency regularization items to imitate the interpolant process in a simple manner: 
\begin{align}
\small \mathcal{L}_{g}&=\mathbb{E}_{p_i \in \mathcal{M}_{e}} \left [ \left \| \Gamma _e(p_i) -\Gamma _e(p_i+\delta  ) \right \| ^2 \right ] , \\
\small \mathcal{L}_{b}&=\mathbb{E}_{p_i \in \partial \mathcal{M}_{e}} \left [ \left \| \Gamma _e(p_i) - \Gamma _b(p_i)\right \| ^2 \right ]  ,  \\
\small \mathcal{L}_{SC} &= \mathcal{L}_{g} + \lambda \mathcal{L}_{b}.
\end{align}
Intuitively, the $\mathcal{L}_{b}$ ensures that the editable texture $\Gamma _e$ is consistent with the base texture $\Gamma _b$ in the adjoining areas $\partial \mathcal{M}_{e}$ as Dirichlet boundary condition, while the $\mathcal{L}_{g}$ extends the consistent style throughout the whole learnable part $\Gamma _e$ with gradient constrain on small noise $\delta$. As shown in Fig.~\ref{fig: ablation}, the $\mathcal{L}_{SC}$ achieves local changes in a seamless manner.


\begin{figure*}[t]
\centering
\includegraphics[width=2.1\columnwidth]{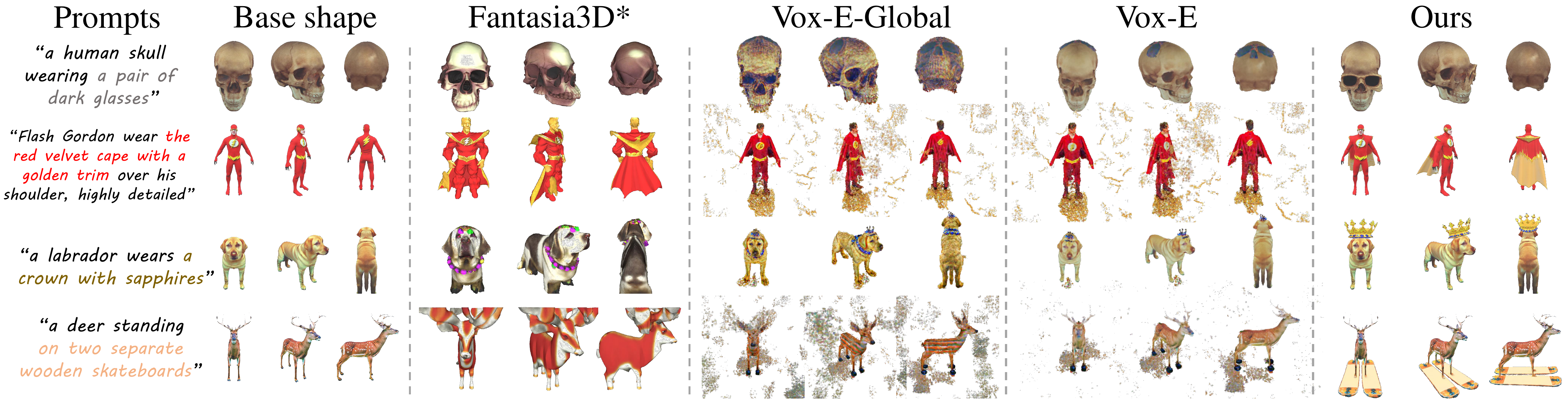} 
\caption{Visual comparison. Our approach synthesizes high-quality edits while preserving the base mesh perfectly.}
\label{fig: 3d}
\end{figure*}

\section{Experiments}
\label{sec: experiment}

\subsection{Experimental Setups}
\noindent \textbf{Implementation Details.} We use the Stable Diffusion implementation by HuggingFace Diffusers for SDS, and adopt DMTet to learn geometry and texture separately with NVDiffRast as a differentiable renderer. FocalDreamer usually takes less than $30$ minutes ($3000$ steps) for geometry and $20$ minutes ($2000$ steps) for texture to converge on $4$ Nvidia RTX 3090 GPUs, where we use AdamW optimizer with the respective learning rates of $1 \times 10^{-3}$ and $1 \times 10^{-2}$. UV edge padding techniques are utilized to remove the seams in the texture maps. More details are provided in the appendix.

\noindent \textbf{Synthetic Object Dataset.} We assemble the dataset with $15$ high-quality meshes found on the internet. We paired each object in our dataset with a detailed edit prompt to showcase our approach's ability to perform \textbf{expressive}, \textbf{precise}, and \textbf{diverse} edits which are absent in other approaches. 

\noindent \textbf{Evaluation Criteria.} Following Vox-E, we report auxiliary quantitative metrics on our dataset: (1) \textit{CLIP Similarity} ($\text{CLIP}_{sim}$) measures the alignment of the performed 3D edits with the text descriptions, and (2) \textit{CLIP Direction Similarity} ($\text{CLIP}_{dir}$) evaluates the edits with the editing directions from the input to edit results, by measuring the directional CLIP similarity between changes of text and 3D shapes, first introduced by StyleGAN-NADA~\cite{gal2022stylegan}.

\noindent \textbf{Baselines.} We compare FocalDreamer with three baselines. (1) \textit{Fantasia3D*}: as claimed in Fantasia3D~\cite{chen2023fantasia3d}, it is able to generate shapes initialized with a low-quality customized 3D mesh. In order to additionally endow it with preservation of texture from base shape, the texture field $\Gamma(p_i)$ is supervised by base texture with reconstruction loss on the base mesh surface, as one of the baselines. 
(2) \textit{Vox-E}~\cite{sella2023vox}: to show our superior editing within desired regions, SOTA editing work Vox-E is also compared. To the best of our knowledge, Vox-E is the only open-source method that directly performs text-guided localized edits for 3D objects.
(3) \textit{Vox-E-Global}: Vox-E also supports global editing to better align with the prompts without constraining from base shape. More details are provided in the appendix.

\begin{figure}[t]
\centering
\includegraphics[width=1\columnwidth]{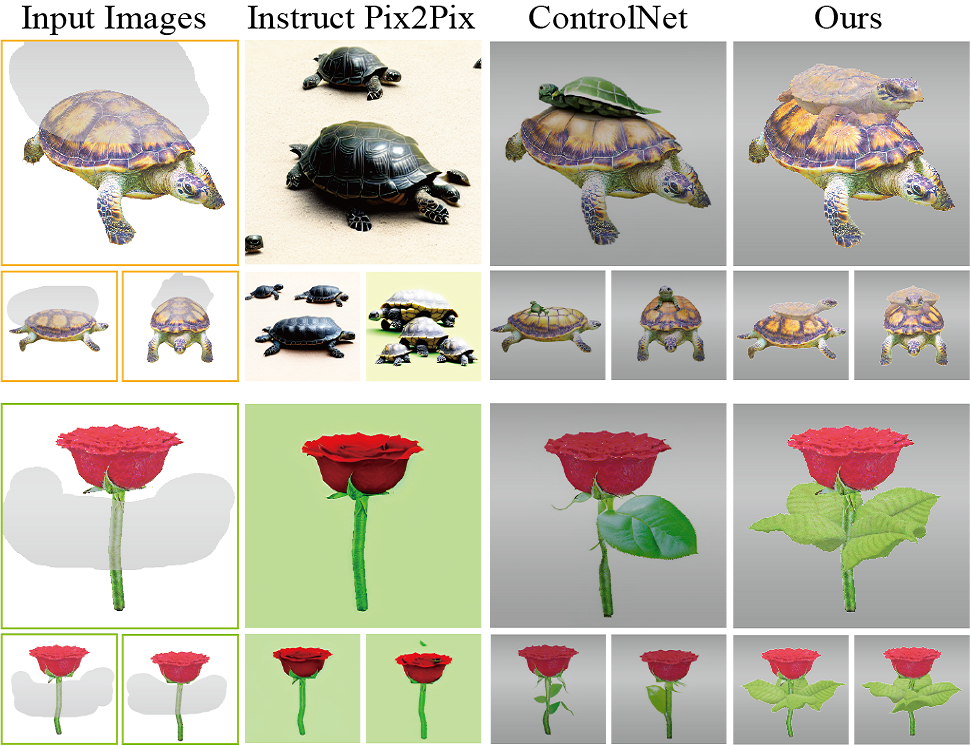} 
\caption{Comparison with SOTA image editing methods. The gray areas in input images indicate the inpainting region specifically added for ControlNet. We observed that 2D editing methods exhibit view-inconsistent, and their editing quality varies notably depending on the viewpoint.}
\label{fig: 2d}
\end{figure}

\begin{table}[t]
\centering
\fontsize{8.5}{9}\selectfont
  \begin{tabular}{lcc}
    \toprule
    Method  & $\text{CLIP}_{sim}$ $\uparrow $  & $\text{CLIP}_{dir}$ $\uparrow $  \\
    \midrule
    Fantasia3D* & 0.284  & 0.0180 \\
    Vox-E-Global & 0.299  & 0.0204 \\
    Vox-E & 0.293  & 0.0178\\
    FocalDreamer (ours) & \textbf{0.329} & \textbf{0.0519} \\
    \bottomrule
  \end{tabular}
    \caption{Quantitative evaluation results across $15$ scenes. }
  \label{tab: main}
\end{table}

\begin{figure}[t]
\centering
\includegraphics[width=1\columnwidth]{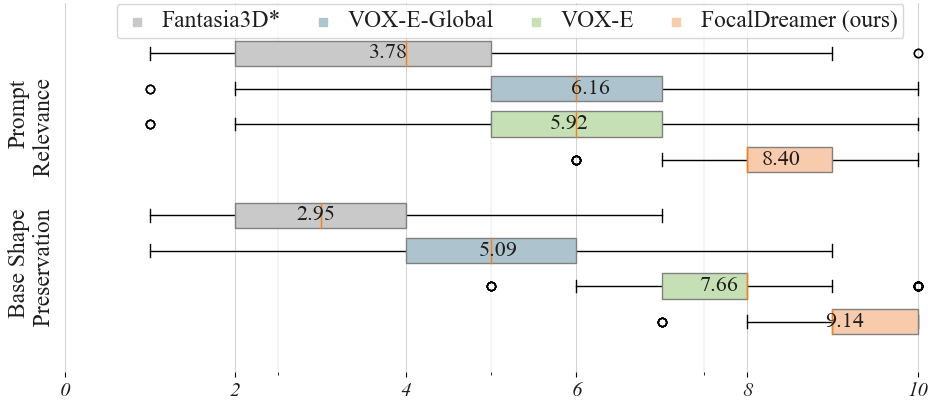} 
\caption{Boxplot illustration of user study. FocalDreamer demonstrates better performance (high means) and stability across scenes (narrow interquartile range).}
\label{fig: user}
\end{figure}

\subsection{Qualitative Results}
The qualitative comparison with 3D editing baselines is shown in Fig.~\ref{fig: 3d} over our dataset. As illustrated in the figure, Fantasia3D* results in an appearance vastly different from the base mesh, even with the texture reconstruction loss, because the whole shape is re-optimized according to prompts. While Vox-E-Global occasionally produces edits that align with prompts, it suffers from subpar editing quality and noticeable outliers. Vox-E demonstrates a limited capacity to filter out undesired changes and noise based on Vox-E-Global, since it heavily relies on a keyword, such as \textit{cape} or \textit{glasses}. Vox-E sometimes misidentifies the focal regions, \textit{i.e.}, placing glasses on the top of the skull. In contrast to them, our editings align perfectly with the prompts while faithfully preserving the details of base mesh, achieving precise and meaningful changes to both geometry and texture.

\noindent \textbf{2D Image Editing Comparisons.} We demonstrate that 2D image editing methods cannot effectively handle 3D object editing tasks. This is primarily because 2D editing on rendered images does not yield satisfactory view-consistent results. We sample renderings from three different viewpoints and apply SOTA image editing methods, namely Instruct Pix2Pix (IP2P)~\cite{brooks2023instructpix2pix} and ControlNet-inpainting (ControlNet)~\cite{zhang2023controlnet}.  We input the same prompts in Fig.~\ref{fig: main} for FocalDreamer and ControlNet, and the modified prompts in the form of ``put xxx on xxx" for IP2P.
As depicted in Fig.~\ref{fig: 2d}, the quality of editing by 2D methods drops significantly from less \textit{canonical} views (\textit{e.g.}, the turtle's left view), and they severely lack view-consistency.

\subsection{Quantitative Results}
We perform a quantitative evaluation in Tab.~\ref{tab: main} on our dataset. To perform a fair comparison, all metrics are calculated with renderings from the same $100$ views across different methods. As illustrated in the table, FocalDreamer achieves noticeably higher $\text{CLIP}_{dir}$. This is attributed to its capability to avoid unnecessary changes and accurately execute the desired editing direction, primarily due to the geometric concentration. Additionally, our editing fidelity ($\text{CLIP}_{sim}$) stands out as the best, stemming from the enhanced part-wise details brought by the separable framework and decoupled learning.

\noindent \textbf{User Study.} While CLIP mainly evaluates the matching degree of rendered views and text prompts, it fails to assess the extent to which the base shape is properly preserved. We conduct user studies with $65$ participants to evaluate different methods based on user preferences across $15$ scenes. We ask the participants to give a preference score (range from $1 \sim 10$) in terms of prompt relevance and base shape preservation for $5$ random views per scene from anonymized methods’ generation. As shown in Fig.~\ref{fig: user}, we report the distribution of the scores, including the medians, means, quartiles and outliers. We find that FocalDreamer is significantly preferred over all baselines in terms of source preservation (\textit{i.e.}, $mean=9.14$) and prompt relevance (\textit{i.e.}, $mean=8.40$). The narrow interquartile range in our method also demonstrates a more stable editing effect across various scenes.

\begin{figure*}[t]
\centering
\includegraphics[width=2.1\columnwidth]{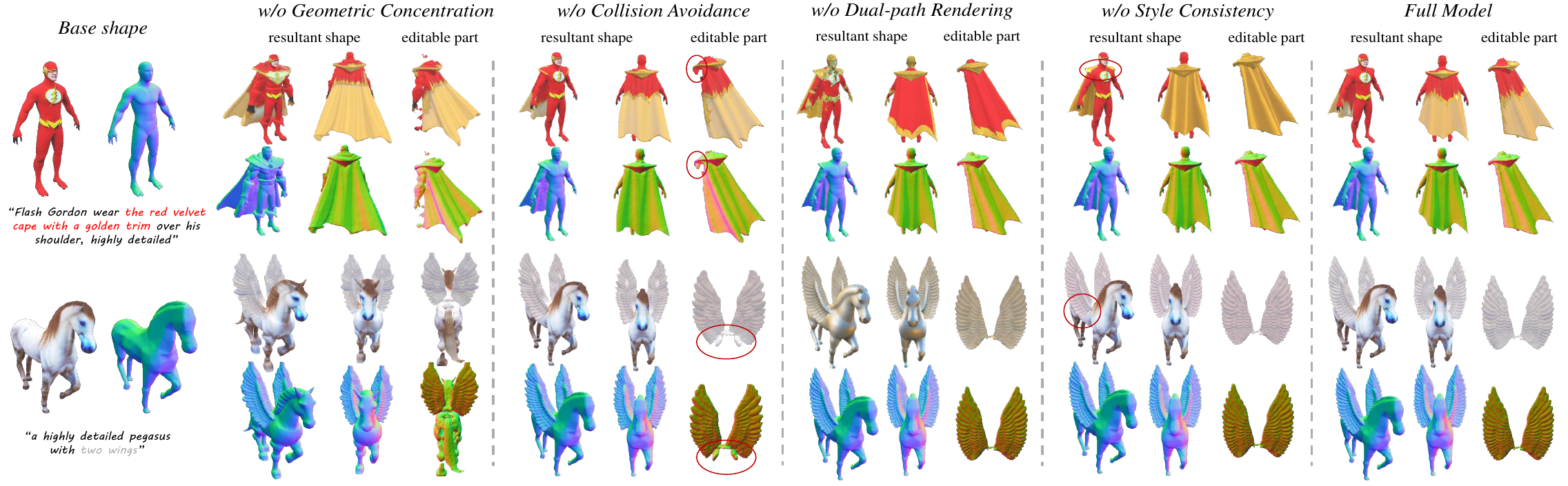} 
\caption{Ablation study. We visually illustrate the effect of each technique we propose. Please refer to Section~\ref{sec: ablation} for details.}
\label{fig: ablation}
\end{figure*}

\begin{figure}[t]
\centering
\includegraphics[width=1\columnwidth]{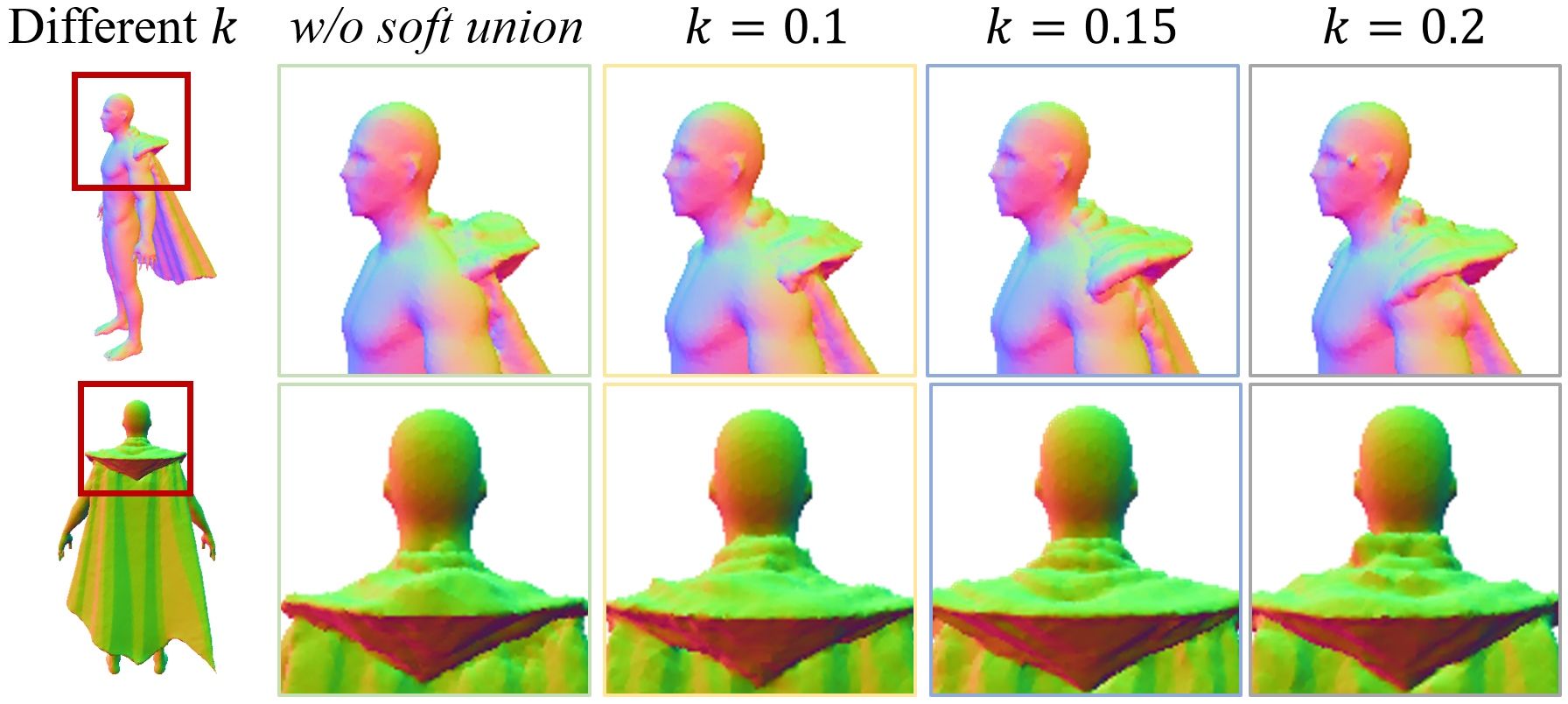} 
\caption{Geometry union sensitivity. The smoothness of the junction varies with different $k$ in Eq. 3 and 4.}
\label{fig: union}
\end{figure}

\begin{figure}[t]
\centering
\includegraphics[width=0.95\columnwidth]{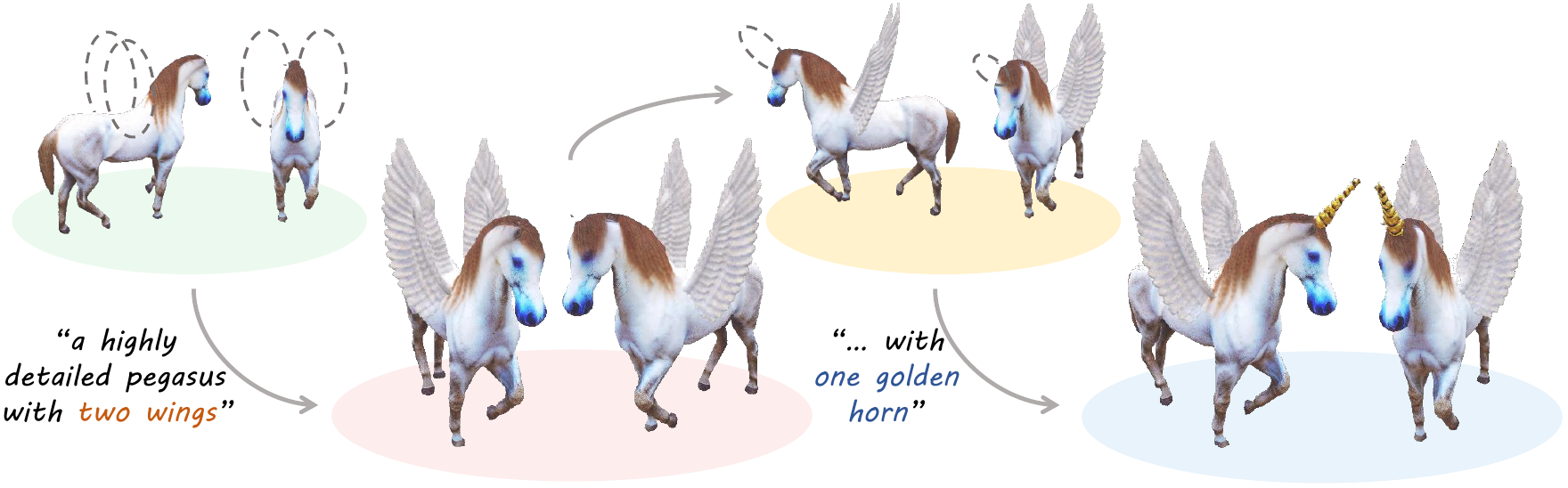} 
\caption{Progressive editing. The horse is first edited by adding two wings, then a horn is added in a subsequent edit.}
\label{fig: progressive}
\end{figure}

\begin{table}[t]
    \centering
    \fontsize{8.5}{9}\selectfont
    \begin{tabularx}{1.0\columnwidth}{ c c c c | c c }  
    \toprule
     $\mathcal{L}_{GF}$ &  $\mathcal{L}_{CA}$  &  $\mathcal{L}_{SC}$  &   Dual-path Render  &  $\text{CLIP}_{sim}$  &  $\text{CLIP}_{dir}$  \\ 
    \addlinespace[3pt]
    \hline
    \addlinespace[3pt]
    \ding{51}    &  \ding{51}   & \ding{55}   &  \ding{51} &   0.312* &  0.0402*    \\ 
    \addlinespace[5pt]
    \ding{51}    &  \ding{51}   & \ding{51}   &  \ding{51}  &   \textbf{0.319}* &  \textbf{0.0495}*    \\ 
    \addlinespace[3pt]
    \hline 
    \addlinespace[3pt]
    \ding{55}    &  \ding{51}   & \ding{51}   &  \ding{51}  &   0.316 &  0.0433    \\
    \ding{51}    &  \ding{55}   & \ding{51}   &  \ding{51}  &   \textbf{0.329} &  0.0517    \\
    \ding{51}    &  \ding{51}   & \ding{51}   &  \ding{55}  &   0.313 &  0.0401    \\ 
    \addlinespace[5pt]
    \ding{51} &  \ding{51}   &  \ding{51}   &  \ding{51}  &  \textbf{0.329}  &  \textbf{0.0519}    \\ 
    \bottomrule
    \end{tabularx}
    \caption{Quantitative ablation study. Since not all scenes require style consistency, we report the metrics of editings require $\mathcal{L}_{SC}$ 
    with $*$. The performance of FocalDreamer would noticeably deteriorate without the above components.}
    \label{tab: ablation}
\end{table}

\subsection{Ablation Study}
\label{sec: ablation}

We conduct the ablation study both qualitatively and quantitatively. In particular, by setting $\mathcal{L}_{GF}$, $\mathcal{L}_{CA}$ and $\mathcal{L}_{SC}$ to zero respectively, we investigated the effects of our proposed \textit{Geometric Concentration}, \textit{Collision Avoidance}, and \textit{Style Consistency} strategies. In order to validate the \textit{dual-path rendering}, we employ the naive idea of single rendering outlined in Section~\ref{sec: texture} for ablation. Specifically, it involves rendering the entire shape with a learnable texture $\Gamma_e$, which is initialized with the base texture $\Gamma_b$.

As illustrated in Fig.~\ref{fig: ablation} and Tab.~\ref{tab: ablation}, $\mathcal{L}_{GF}$ significantly constrains geometric alterations outside the focal region, resulting in localized edits. $\mathcal{L}_{CA}$ effectively prevents undesirable geometric overlap within the base mesh, especially at the junction like the root of \textit{wings} and \textit{capes}. Since $\mathcal{L}_{CA}$ predominantly affects the purity of the editable part and has minimal impact on the overall appearance, its quantitative metrics closely align with the full model. In the absence of \textit{dual-path rendering}, the base mesh texture experiences unintended alterations due to the update of the whole texture network during appearance learning. Moreover, editing with $\mathcal{L}_{SC}$ exhibits a harmonious overall style and nature transition in certain instances, but it is not universally required (\textit{e.g.}, a butterfly over a tree stump). In Tab.~\ref{tab: ablation}, we use $*$ to denote scenes that require $\mathcal{L}_{SC}$ for a fair comparison.

\noindent \textbf{Progressive Editing.} Our method can be used as a sequential editor for users' requirements, and progressively edits base mesh. In Fig.~\ref{fig: progressive}, we exhibit a two-step editing by first generating \textit{two wings} on \textit{horse}, followed by adding \textit{a horn}.

\noindent \textbf{Geometry Union Sensitivity.} We also demonstrate the smoothness of the junction between the editable part and base mesh with various $k$ (Eq. 3 and 4) in Fig.~\ref{fig: union}. It is evident that larger $k$ leads to a more natural but pronounced transition region. We set $k=0.15$ for a moderate transition.

\section{Conclusion}
In this paper, we present FocalDreamer, a text-driven framework that supports separable, precise, and consistent local editing for 3D objects. Technically, we equipped FocalDreamer with geometry union and dual-path rendering to assemble independent 3D parts, facilitating instance reuse and part-wise control. Geometric focal loss and style consistency regularization are proposed to encourage focal fusion and congruent overall appearance. Comprehensive experiments and detailed ablation studies have demonstrated our approach possesses superior local editing power through a well-conceived framework design. We hope that FocalDreamer will help pave the way for expressive, localized 3D content editing for casual artists, bringing us closer to the goal of democratizing 3D content creation for all.

{\small
\bibliographystyle{ieee_fullname}
\bibliography{egbib}
}

\end{document}


\title{Supplementary Materials:   

Generalized Deep 3D Shape Prior via Part-Discretized Diffusion Process
}

\author{Yuhan Li\textsuperscript{1}\hspace{2mm}
Yishun Dou\textsuperscript{2}\hspace{2mm}
Xuanhong Chen\textsuperscript{1}\hspace{2mm}
Bingbing Ni\textsuperscript{1, 2$\dagger$}\hspace{2mm}
Yilin Sun\textsuperscript{1}\hspace{2mm}
Yutian Liu\textsuperscript{1}\hspace{2mm}
Fuzhen Wang\textsuperscript{1}\\
\textsuperscript{1}Shanghai Jiao Tong University, Shanghai 200240, China \qquad \textsuperscript{2}Huawei \\
{\tt\small \{melodious, nibingbing\}@sjtu.edu.cn}\\
{\small \url{https://github.com/colorful-liyu/3DQD}}
}

\maketitle

\newcommand{\customfootnotetext}[2]{{
  \renewcommand{\thefootnote}{#1}
  \footnotetext[0]{#2}}}
\customfootnotetext{${\dagger}$}{Corresponding author: Bingbing Ni.}


In Section~\ref{sec: limit}, we provide the limitations of our work and and future work. Building upon experimental insights, we delve into the stability and sensitivity issues of the CLIP model in Section~\ref{sec: clip}. In Section~\ref{sec: baseline}, we provide additional details about the baselines, along with an extensive discussion of two potential related methods (DreamEditor~\cite{zhuang2023dreameditor} and Instruct-NeRF2NeRF~\cite{haque2023instruct}). The evaluation metrics and user study are elaborated upon in Section~\ref{sec: expe}. In Section~\ref{sec: imple}, we release the implementation details of FocalDreamer across its various stages. Lastly, we introduce the contents of two additional files in the supplementary materials in Section~\ref{sec: zip}. Once the paper is accepted, we will open source our complete code on GitHub.

\begin{figure*}[t]
\centering
\includegraphics[width=2\columnwidth]{cvpr2023-author_kit-v1_1-1/image/overlay.pdf}
\caption{We showcase the scope of our work's editing capabilities, which encompass additive and overlay 3D editing.}
\label{fig: overlay}
\end{figure*}

\begin{figure}[t]
\centering
\includegraphics[width=1\columnwidth]{cvpr2023-author_kit-v1_1-1/image/clip.pdf}
\caption{We visually illustrate the impact of different texts on CLIP similarity calculations using three examples. On the left side, we present the rendering results, while on the right side, blue text represents the original complex prompts, and orange text signifies the simplified prompts. Additionally, we report the average $\text{CLIP}_{sim}$ for each example based on $100$ rendering results.}
\label{fig: clip}
\end{figure}

\section{Limitations and Future Work}
\label{sec: limit}
We present FocalDreamer, a text-driven framework that supports separable, precise, and consistent local editing for 3D objects. However, there are several limitations to discuss, which will help us improve the proposed framework further.

First, since we propose the separable pipeline that combines base shape with learnable parts, we are only able to achieve additive and overlay edits as shown in Fig.~\ref{fig: overlay}. Removing portions of the base mesh is not supported.

Moreover, similar to previous works~\cite{chen2023fantasia3d, poole2022dreamfusion, mikaeili2023sked}, our approach utilizes SDS Loss and may be vulnerable to the well-known ``multiface issue" depending on the choice of diffusion model and prompt. This could stem from the 2D supervision from SDS. 2D supervision results in a lack of viewpoint information, which in turn leads to the absence of multi-view semantic alignment in 3D generation~\cite{poole2022dreamfusion}. It causes optimization results to vary semantically across different views, such as ``multiface issue". 

Finally, our results rely on the publicly available Stable-Diffusion model~\cite{rombach2022ldm}, which is less amenable to directional text prompts and produces lower quality 3D generated outputs compared to commercial diffusion models used by previous works~\cite{lin2023magic3d, poole2022dreamfusion}.

Future directions may expand our method to remove portions of the base mesh. Another potential direction is to explore the way to import viewpoint information into distilling for more aligned views semantically.

\begin{table}[t]
\centering
\fontsize{9}{10}\selectfont
  \begin{tabular}{lcc}
    \toprule
    Method  & $\text{CLIP}_{sim}$ $\uparrow $  & $\text{CLIP}_{dir}$ $\uparrow $  \\
    \midrule
    FocalDreamer (complex) & 0.329  & 0.0519\\
    FocalDreamer (simplified) & 0.336 & 0.0538 \\
    \bottomrule
  \end{tabular}
    \caption{We show the impact of different texts on CLIP scores quantitatively. The simplified texts obtain high scores.}
  \label{tab: simplify}
\end{table}

\section{CLIP Evaluation Analysis}
\label{sec: clip}
We notice that CLIP similarity $\text{CLIP}_{sim}$ values vary significantly across different scenes. For instance, Vox-E achieves a $\text{CLIP}_{sim}$ mean score of $0.36$ on their dataset (with instances like ``wearing sunglasses," ``a wizard’s hat," and ``a Christmas sweater"). However, the $\text{CLIP}_{sim}$ mean score drops to only 0.293 when we apply Vox-E on our dataset. Intrigued by this observation, we conducted an in-depth analysis.

We identified a significant distinction between our dataset and Vox-E's dataset, mainly centered around the complexity of the editing prompts. Their prompts consist of fewer than ten words, providing concise descriptions of the editing subject and the editable attributes (``A 〈object〉 wearing sunglasses"). Conversely, our dataset contains detailed descriptions and expressive adjectives (``a deer standing on two separate wooden skateboards," ``a highly detailed pegasus with two wings"). We hypothesize that the capability of the CLIP model is too limited to understand such complex concepts clearly due to both CLIP's training dataset and its scale. Consequently, it may not capture the nuances of intricate prompts, leading to adverse effects on similarity scores for more complex prompts during evaluation.

\begin{figure}[t]
\centering
\includegraphics[width=1\columnwidth]{cvpr2023-author_kit-v1_1-1/image/in2n.png}
\caption{The original images from the Instruct-N2N paper~\cite{haque2023instruct}. They provide a visual representation of the limitations of Instruct-N2N's capabilities:(1) the inability to perform large spatial manipulations, (2)
adding or removing large objects. They are the primary reasons behind its failure in our task.}
\label{fig: in2n}
\end{figure}

To verify our hypothesis, we first edit 3D shapes on our dataset using FocalDreamer with complex prompts. Next, we conduct an evaluation by creating two sets of text prompts with corresponding $100$ rendering images into the CLIP model. One set consisted of the original complex prompts used during editing, while the other set involved simplified prompts, retaining only the core of the sentences. We present the scores for these two sets of text prompts in Tab.~\ref{tab: simplify} and provide visualizations in Fig.~\ref{fig: clip}. Notably, we find that our model's scores on 15 scenes improve significantly when simplified prompts are provided to CLIP, confirming our hypothesis. This suggests that CLIP model is highly sensitive to scene and text formats, and its scoring capabilities may not be entirely stable. Furthermore, the disparity in Vox-E's performance on our dataset compared to theirs is well justified.

\section{Baseline Discussions}
\label{sec: baseline}
\subsection{Baselines Compared in Main Paper}
\subsubsection{Fantasia3D*.} When comparing to Fantasia3D we used the code provided by the authors. However, the official codes support only generation based on an untextured mesh. To preserve the texture of base mesh for a fair comparison, we add MSE loss between the base mesh texture and edited mesh texture in 3D space.

\subsubsection{Vox-E and Vox-E-Global.} We used the official code provided by the authors.

\subsection{Other Baselines}
We discuss some other potential baselines (except Fantasia3D and Vox-E) related to our work in this section.
\subsubsection{DreamEditor.} By representing scenes as mesh-based neural fields, DreamEditor~\cite{zhuang2023dreameditor} allows localized editing within specific regions with the attention map to identify the regions to be edited. Actually, it uses the same method to localize as Vox-E. Because DreamEditor authors \textit{haven't  open-sourced the codes}, we cannot compare it quantitatively and qualitatively. From the results released in their paper, we find DreamEditor only supports minor shape insertion and replacement of objects of the same type (\textit{e.g.}, replacing a horse with a deer, inserting a small rose near the mouse of a dog), which is due to its direct operating on the level of mesh vertices and predicting the attributes of vertices. However, the number of vertices cannot be changed, which limits its geometry modification significantly. Moreover, it doesn’t achieve separable editing which is vital for instance reuse and part-wise control. As reported by the authors of DreamEditor, DreamEditor does not directly model environmental lighting, which limits control over the lighting condition. And it also includes the ``multiface problem".

\subsubsection{Instruct-NeRF2NeRF.} Given a NeRF of a scene and the collection of images used to reconstruct it, Instruct-NeRF2NeRF uses InstructPix2Pix~\cite{brooks2023instructpix2pix} to iteratively edit the input images while optimizing the underlying scene, resulting in a \textbf{global} 3D editing.

We evaluate Instruct-NeRF2NeRF on our task, but we did not observe any successful edits. When provided with editing text prompts, the base scenes do not undergo edits that match the given text; instead, they generate blurred reconstruction results. This situation aligns with the limitations reported by Instruct-NeRF2NeRF itself in their paper: ``(1) the inability to perform large spatial manipulations, (2) adding or removing large objects, (3) certain views may consistently produce less of an edit or no edit at all.", as shown in the Fig.~\ref{fig: in2n} released in their official paper.

In fact, all the reported results in their paper lean toward texture and style edits, with minimal attention on substantial geometric manipulations, which is important for our task. This outcome could stem from (1) the coupling of texture and geometry learning, resulting in a lack of significant geometric variations. The model might easily become trapped in optimizing local texture optima. (2) Furthermore, geometric alterations lead to dataset images with significant multi-view disparities, which isn't conducive to iterative dataset update-based editing in Instruct-NeRF2NeRF. (3) Additionally, Instruct-NeRF2NeRF's editing capability is rooted in the 2D large-scale model Instruct pix2pix~\cite{brooks2023instructpix2pix}. But Instruct pix2pix has not succeeded in achieving the desired edits we have illustrated in the image comparison figure in the main paper. These $3$ potential reasons collectively contribute to the complete failure of Instruct-NeRF2NeRF on our task. Thus, we have chosen not to include Instruct-NeRF2NeRF as one of the baselines in our main text.

\section{Experiments}
\label{sec: expe}
\subsection{Evaluation Protocol}
Due to space constraints, our introduction of the evaluation metrics in the main text is brief. In the following, we elaborate on the metrics we adopt in detail. The evaluation metrics are the same in Vox-E~\cite{sella2023vox}.
 \smallskip \newline \emph{CLIP Similarity} ($\text{CLIP}_{sim}$) measures the semantic similarity between the output objects and the target text prompts. We encode both the prompt and images rendered from our 3D outputs using CLIP's text and image encoders, respectively, and measure the cosine distance between these encodings. 
 \smallskip \newline \emph{CLIP Direction Similarity} ($\text{CLIP}_{dir}$) evaluates the quality of the edit in regards to the input by measuring the directional CLIP similarity first introduced by Gal et al.~\cite{gal2022stylegan}. This metric measures the cosine distance between the direction of the change from the input and output rendered images and the direction of the change from an input prompt (\emph{i.e.} ``a highly detailed white horse") to the one describing the edit (\emph{i.e.} ``a highly detailed pegasus with two wings").

To evaluate our results quantitatively, all $15$ scenes in our Synthetic Object Dataset are used, with the prompts offered in the \textit{FocalDreamer-edit-gallery.zip} file. Please refer Section~\ref{gallery} for more details. During the evaluation, we render each edited scene from 100 different poses distributed evenly along a $360^{\circ}$ ring at the elevation angle $\theta =\frac{\pi}{4}$. In addition to these $15$ scenes, we also render 100 images from the same poses on the base shape for each input scene. When comparing our result with other 3D textual editing papers we evaluate our results using two CLIP-based metrics. %
The CLIP model we used for both of these metrics is ViT-B/32 and the base text prompts used to calculate the directional CLIP metric are also included in \textit{FocalDreamer-edit-gallery.zip}.

\subsection{User Study Details} 
We engaged $65$ participants in the user study, wherein they assessed $15\times 4$ meshes edited by four distinct methods, all derived from the same textual prompt. We show $5$ rendered images for each mesh at the elevation angle $\theta =\frac{\pi}{4}$. The base shape is also shown in $5$ images with the same angles. Participants were prompted with two inquiries: (Q1) ``To what extent does the outcome align with the text description?" (Q2) ``To what extent is the base shape preserved?" Then participants are asked to give a preference score (range from $1 \sim 10$) for the above two questions. The distribution of the scores is reported in the paper's main body. We find that FocalDreamer is significantly preferred over all baselines in terms of source preservation (\textit{i.e.}, $mean=9.14$) and prompt relevance (\textit{i.e.}, $mean=8.40$). The narrow interquartile range in our method also demonstrates a more stable editing effect across various scenes.

\begin{figure}[t]
\centering
\includegraphics[width=1\columnwidth]{cvpr2023-author_kit-v1_1-1/image/geometry.pdf}
\caption{We demonstrate how the rendering results are optimized by the diffusion model during the geometry learning stage. The geometry learning stage comprises two phases: coarse and refined.}
\label{fig: geometry}
\vspace{-3mm}
\end{figure}

\begin{figure}[t]
\centering
\includegraphics[width=1\columnwidth]{cvpr2023-author_kit-v1_1-1/image/offsets.pdf}
\caption{Illustration of the effect of the 3D mask for grid offsets. Global vertex offsets result in subtle changes to the base mesh geometry.}
\label{fig: offset}
\end{figure}

\section{Implementation}
\label{sec: imple}
\subsection{Geometry Learning}
During geometry learning, we aim to optimize the hash-grid-based~\cite{muller2022instant} MLP with parameters $\Psi _e$. We implement the $\Psi_e$ as a two-layer MLP with 32 hidden units. Specifically, as shown in Fig.~\ref{fig: geometry}, geometry learning is divided into two sub-phase: coarse and refined. The first $\frac{2}{3}$ steps for geometry learning are coarse phase, while the rest are refined phase. 

During the coarse phase, we utilize a differentiable renderer to obtain a normal map $n$ and an object mask $o$. Subsequently, $n$ and $o$ are directly concatenated (noted as $\tilde{n}$), and downsampled to the same size ($64\times 64\times 4$) as the latent codes $z^{\tilde{n}} $ of the diffusion model, serving as the input of shape encoding to stable diffusion. This is inspired by Latent-NeRF~\cite{metzer2023latentnerf}, for it helps achieve better convergence and speed up the training process. 

In the refined phase, the object mask $o$ is discarded, and we directly encode the normal map $n$ using the pre-trained encoder from large T2I models to derive $z^{\tilde{n}} $. Through the encoder, high-resolution normal images facilitate capturing finer geometry details compared with the coarse phase. We observed that the coarse phase is very crucial for the overall shape of the editable part during the editing~\cite{tang2023make, chen2023fantasia3d}. Since the refined phase mainly focuses on attaining finer geometry details, no notable geometry structural change is observed. 

\subsubsection{Local Vertex Offsets.} Additionally, we observe the vertex offset predicted by $\Psi _e$ for $v_i$ may cause undesired changes on base mesh's surface. To tackle this, we apply a 3D mask constraint ($s_e\ge  0$) so that only vertices close to the learnable shape can deform, which helps in preserving the geometric appearance of the base shape accurately.  In Fig.~\ref{fig: offset}, we visualize the result of using vertex offset for all deformable tetrahedral vertices causing a slight base mesh change, while this problem is solved by the 3D mask.

\subsubsection{Hyper-parameters.} The SDS loss, geometric focal loss $\mathcal{L} _{GF}$, and collision avoidance $\mathcal{L} _{CA}$ loss are used during geometry learning:
\begin{equation}
\mathcal{L} _{geo} = \mathcal{L} _{SDS} + \lambda _{GF} \mathcal{L} _{GF} + \lambda _{CA} \mathcal{L} _{CA},
\end{equation}
where $\lambda _{GF}=1000$ and $\lambda _{CA}=100$ as default.

\subsection{Appearance Learning}
After the optimization of the geometry network, the resultant mesh $\mathcal{M}$ is obtained from the soft fusion and MT layer. We design the dual-path rendering which has been explained in the main paper. However, the rendering details for each branch are omitted for the space limitations. We now introduce the PBR material used in each branch in this section.

\subsubsection{PBR Material.}
Following the Physically Based Rendering (PBR) material model, each texture neural field comprises three components, namely diffuse term $k_d \in \mathbb{R}^3$, the roughness and metallic term $k_{rm} \in \mathbb{R}^2$, and the normal term $k_n \in \mathbb{R}^3$. As a result, the specular lobe is described by the specular highlight color:
\begin{equation}
k_s=(1-m)\cdot 0.04+m\cdot k_d , 
\end{equation}
where $m$ represents the metalness parameter. For any point $p\in \mathbb{R}^3$ in the space, we use hash-grid encoding~\cite{muller2022instant} MLP $\Gamma$ to parameterize the texture fields as:
\begin{equation}
(k_d, k_{rm},k_n)=\Gamma (\beta(p)),
\end{equation}
where $\beta$ is the positional encoding of $p$. The MLP $\Gamma$ has only a single-layer with 32 hidden units. At last, each image pixel at a specific viewing direction can be rendered according to the rendering equation~\cite{kajiya1986rendering} as 
\begin{equation}
L(\omega _o)=\int_{\Omega } L_i(\omega _i)f(\omega _i, \omega _o)(\omega _i\cdot n)d\omega _i.
\end{equation}
This is an integral of the product of the incident radiance, $L_i(\omega _i)$ from direction $\omega _i$ and the BSDF $f(\omega _i, \omega _o)$.  The outgoing radiance $L(\omega _o)$ consists of two parts: specular intensity and diffuse intensity. Please refer to NVdiffRec~\cite{munkberg2022dmtet} for more details.

\subsubsection{Hyper-parameters.} The SDS loss, and style consistency loss $\mathcal{L} _{SC}$ loss are used during appearance learning:
\begin{align}
\mathcal{L} _{SC} &= \mathcal{L} _{g} + \lambda_{b} \mathcal{L} _{b}\\
\mathcal{L} _{appear} &= \mathcal{L} _{SDS} + \lambda _{SC} \mathcal{L} _{SC},
\end{align}
where $\lambda _{b}=100$ and $\lambda _{SC}=10$ as default for the scenes need coherent surface.

\subsection{Camera and Light Settings}
\subsubsection{Cameras.} Following Fantasia3D~\cite{chen2023fantasia3d}, we determine camera positions using the spherical coordinate system, characterized by ($r$, $\theta $, $\varphi $). Here, $r$ signifies the radius, $\theta $ stands for the elevation, and $\varphi $ represents the azimuth angle. Specifically, for a batch of $s \times l$ images, we divide the entire azimuth angle range $\varphi $, which spans from $[0, 2 \pi]$, into l segments. Each segment has an interval of $[\frac{2d\pi }{l}, \frac{2(d+1)\pi }{l}]$. Within each segment, we uniformly select s azimuth angles. Consequently, this stipulates the condition for the k-th viewpoint:
\begin{equation}
\left\{\begin{matrix}
r_k \in & [r_{min}, r_{max}], \\
\theta_k \in &[\theta _{min}, \theta_{max}, \\
\varphi _k \in &[\frac{2d\pi}{l},\frac{2(d+1)\pi}{l}),
\end{matrix}\right.
\end{equation}
where $r_{min}, r_{max}, \theta_{min}, \theta_{max}$ are hyperparameters that decide the range of $r$ and $\theta$, and $d=k$ mod $l$. In all cases, we set $r_{min}=r_{max}=3$, $\theta_{min}=-\frac{\pi}{18}$ and $\theta_{max}=\frac{\pi}{4}$. During sampling to evaluate, the elevation angle of $\theta $ is set to be $\theta_{max}=\frac{\pi}{4}$ for all.

Besides, the direction description about the azimuth angle $\varphi $ is added on the editing prompts, in the form of ``\{editing prompts\}, 〈XXX〉 view". The ``〈XXX〉" view is selected from ``front", ``side", ``back".

\subsubsection{Light.} We adopt fixed HDR environment maps as light. Learning the PBR materials is an ill-posed problem. If materials and lighting are learned together, it will increase the difficulty of learning~\cite{chen2023fantasia3d}. So We use the fixed HDR light to optimize the appearance. 

\subsection{Focal Region Initialization}
To offer more precise control over the generation process, users are requested to select one or multiple ellipsoid areas (depending on the editing needs) as focal/target regions. Each focal region $\Omega '$ is deformed from a standard sphere $\Omega $ by an affine transformation with $9$ degrees of freedom (DOF), $3$ DOF for stretching, $3$ DOF for rotation, and $3$ DOF for translation along the \{X, Y, Z\}\text{-axis}:
\begin{equation}
\small \Omega '=R_{xyz}(\alpha, \beta, \gamma)\cdot T(t_x, t_y, t_z)\cdot S(s_x, s_y, s_z)\cdot \Omega .
\end{equation}

To be more precise, we provide the detail of the affine transformation to deform the sphere $\Omega $ to ellipsoid $\Omega '$ as follows:
\begin{equation}
\small \begin{bmatrix}
 x'\\
 y'\\
 z'\\
1
\end{bmatrix}
=
\begin{pmatrix}
  1&  &  &t_x \\
  &  1&  &t_y \\
  &  &  1&t_z \\
  &  &  & 1
\end{pmatrix}
\cdot
\begin{pmatrix}
  s_x&  &  & \\
  &  s_y&  & \\
  &  &  s_z& \\
  &  &  & 1
\end{pmatrix} 
R_{xyz}(\alpha, \beta, \gamma)\cdot
\begin{bmatrix}
 x\\
 y\\
 z\\
1
\end{bmatrix},
\label{affine}
\end{equation}

where the [x', y', z'] is the sampling point on the surface of ellipsoid $\Omega '$, and the [x, y, z] is the point on the surface of standard sphere $\Omega $. $R_{xyz}$ represents the rotation matrix along the \{X, Y, Z\}\text{-axis} and can be simplified according to Rodrigues' rotation formula~\cite{rotation}.

In terms of the actual code implementation, our pipeline for focal regions is as follows: (1) We import a standard sphere mesh into FocalDreamer as $\Omega $, and then the affine transformation in Eq.~\ref{affine} is applied to deform the sphere into the desired region. (2) During the initialization of DMTet~\cite{munkberg2022dmtet}, we employ the CUBVH~\cite{cubvh} to calculate the signed distance function (SDF) for points in the whole space. (3) We sample a batch of $10240$ points each time with their SDF values to initialize DMTet. The initialization process, which includes sampling the SDF and performing a coarse fitting, typically takes approximately two minutes ($15000$ iterations).

\subsection{Score Distillation Sampling Settings}
We employ a slightly different SDS loss between geometry learning and appearance learning, adjusting the maximum time-step from which we derive $t$. Specifically, the SDS loss incorporates two supplementary hyper-parameters into our framework: a starting time-step $t_0$, and an ending time-step $t_{final}$. 
\begin{equation}
    t \sim \mathcal{U}[t_0 , t_{final}].
\end{equation}
Noise is then added to the rendered image according to the time-step sampled from the fitting distribution. During geometry learning, $t_0=0.02$ and $t_{final}=0.35$ in most cases.
And the $t_0$ keeps the same but $t_{final}=0.98$ during appearance learning. 
It's mainly because we find larger $t_{final}$ usually promotes more parts ``growing" based on the base mesh. To keep the geometry generation stable, moderate $t_{final}$ is used for geometry learning.


\section{Introduction for the \textit{zip} file submitted}
\label{sec: zip}
\subsection{FocalDreamer-edit-gallery.zip} 
\label{gallery}
Due to the file size limitation of small than $50$ MB in the AAAI submission system, we were unable to upload all of our result files. Therefore, within the \textit{FocalDreamer-edit-gallery.zip}, we have submitted most of our editing results in the form of GIF animations. Files with the suffix ``base" indicate the base shape, while those with ``edit" denote the edited results. Within the \textit{prompts} folder, all editing text prompts are included, as well as the base prompts that describe the base mesh. The latter is primarily used when calculating $\text{CLIP}_{dir}$.

\subsection{supp-for-video-demo-and-dataset.zip} 

Due to the file size limitation of less than 50MB in the AAAI system, we only uploaded a representative sample from our Synthetic Object Dataset. This sample includes its topology files as well as $k_n$, $k_s$, $k_d$, and lighting HDR. The complete Synthetic Object Dataset comprises 15 such shapes.

In addition, throughout the training process, we continuously sampled the results, generating a video file located in the \textit{training-process-video-demo} folder. This video demonstrates the geometric and textural changes of the \textit{rose} and \textit{turtle} examples as the number of training iterations increases. We have also included the training results from Vox-E~\cite{sella2023vox} for readers to review. The \textit{baseshape-init.mp4} showcases the initialization results of the base shape in Vox-E. It's evident that the base shape initialization is successful and noise-free. The \textit{edit-global.mp4} and \textit{edit-local.mp4} videos display the global and local editing results using Vox-E, respectively. Notably, in the \textit{flash} demo, Vox-E, being unable to precisely pinpoint the focal editing region, results in a significant amount of noise generation, particularly near the {Flash's feet}. Moreover, when Vox-E employs Spatial Refinement via 3D Cross-Attention, it evidently fails to accurately locate the attention area of the \textit{cape} and filter out irrelevant edits.

In the \textit{evluation-scripts} folder, we've released the test scripts for a more intuitive understanding of our evaluation metrics. Readers can utilize this script by following these steps: (1) Decompose the edited results' GIF animations from the \textit{FocalDreamer-edit-gallery} folder into individual images and save them in the \textit{validate} folder. Similarly, decompose the base shape GIF animations and save them in the \textit{validate-baseshape} folder. (2) In the parent directory, save the base text prompts and editing text prompts in \textit{prompt-base.txt} and \textit{prompt.txt} respectively. (3) Execute the script and use the ``-d" parameter to point to the aforementioned locations.

\begin{figure*}[t]
\centering
\includegraphics[width=1.7\columnwidth]{cvpr2023-author_kit-v1_1-1/image/3Dcompare_supp.pdf}
\caption{More comparison with 3D editing methods.}
\label{fig: morecompare}
\end{figure*}

\clearpage
{\small
\bibliographystyle{ieee_fullname}
\bibliography{egbib}
}